\pdfoutput=1
\documentclass[11pt]{article}

\usepackage[]{acl}

\usepackage{times}
\usepackage{latexsym}
\usepackage[normalem]{ulem}

\usepackage[T1]{fontenc}
\usepackage[utf8]{inputenc}

\usepackage[]{xcolor}
\usepackage{microtype}
\usepackage{graphicx}
\usepackage{subcaption}
\usepackage{booktabs}
\usepackage{amsmath}
\usepackage[inline]{enumitem}
\usepackage{tablefootnote}
\usepackage{pgfplotstable} 
\usetikzlibrary{arrows}
\usepackage{tabularx}
\usepackage{stackengine}
\stackMath
\usepackage{graphicx}
\newcommand{\name}[1]{\textsc{HEAR}%\includegraphics[scale=0.08]{pear.png}
}
\newcommand{\encoder}[1]{\textsc{Hybrid} encoder}
\newcommand{\adapter}[1]{\textsc{ARM}}
\newcommand{\unidirec}[1]{\overset{\text{\scriptsize$\rightarrow$}}{#1}}
\newcommand{\bidirec}[1]{\overset{\text{\scriptsize$\leftrightarrow$}}{#1}}
\newcommand{\bH}{\mathbf{H}}
\newcommand{\BY}{\mathbf{Y}}
\newcommand{\bY}{\mathbf{y}}
\newcommand{\bX}{\mathbf{x}}
\newcolumntype{Y}{>{\centering\arraybackslash}X}
\definecolor{diagramblue}{RGB}{61,133,198}
\definecolor{diagramorange}{RGB}{230,145,56}
\definecolor{diagramlightorange}{RGB}{244,227,205}
\definecolor{diagrampurple}{RGB}{195,120,160}
\definecolor{diagramgreen}{RGB}{149,196,121}

\title{Efficient Encoders for Streaming Sequence Tagging}

\author{Ayush Kaushal$^\clubsuit$${\thanks{  ~~Work done as part of an internship at Google.}}$~~~~~~Aditya Gupta$^\spadesuit$~~~~~~Shyam Upadhyay$^\spadesuit$~~~~~~Manaal Faruqui$^\spadesuit$ \\
$^\spadesuit$Google Assistant \\
$^\clubsuit$The University of Texas at Austin \\
{\tt ayushk4@utexas.edu,  \{gaditya, shyamupa, mfaruqui\}@google.com}
  }

\begin{document}
\maketitle
\begin{abstract}

A naive application of state-of-the-art bidirectional encoders for streaming sequence tagging would require re-encoding all tokens from scratch whenever a new token appears in an incremental streaming input (like transcribed speech). The lack of re-usability of previous computation leads to a higher number of Floating Point Operations (or FLOPs) and higher number of unnecessary label flips. Increased FLOPs consequently lead to higher wall-clock time and increased label flipping leads to poorer streaming performance. In this work, we present \textbf{H}ybrid \textbf{E}ncoder with \textbf{A}daptive \textbf{R}estart (\name{}) that addresses these issues while maintaining the performance of bidirectional encoders over offline (or complete) inputs and improving performance on streaming (or incomplete) inputs. \name~ uses a \textsc{Hybrid} unidirectional-bidirectional encoder architecture to perform sequence tagging, along with an Adaptive Restart Module (\adapter~) to selectively guide the restart of bidirectional portion of the encoder. Across four sequence tagging tasks, \name~ offers FLOPs savings in streaming settings upto 71.1\% and also outperforms bidirectional encoders for streaming predictions by upto +10\% streaming exact match.
\end{abstract}

\section{Introduction}
State-of-the-art text encoding methods assume the \textit{offline setting}, where the entire input text is available when encoding it. This differs from the \textit{streaming setting} where the input text grows over time (such as transcribed speech or a typed query) \cite{neural_smt_2016, neural_smt_2017,streaming_query_processing}. Processing streaming input incrementally can enable suggestions on partial inputs \cite{from_simultaneous_to_streaming_machine_translation}, reduced final latency \cite{online_semantic_parsing_acl22} and lead to more interactive NLU agents \cite{query_auto_completion}.% Streaming processing is also fundamental to the way humans process spoken words or written text \cite{temporal_structure_marslen_wilson_and_tyler_1980_elsevier, william_marslen_wilson_1975_science}.

\begin{table}[t]
\small
\setlength{\tabcolsep}{2.5pt}
\centering\begin{tabular}{l  c  c  c}
    \toprule
    Model       & GFLOPs   & Offline F1  & Streaming EM \\
    \midrule
    BiDi encoder        & 74.7      & \textbf{93.1}       & 75.1 \\
    \encoder~         & 43.4      & 93.0       & 72.5 \\
    % BiDi+\adapter & 34.9      & 93.1       & 82.8 \\
    \name~ {\scriptsize(\textsc{Hybrid} + \adapter~)}      & \textbf{21.6}      & 93.0       & \textbf{85.1} \\
    % UniDi        & 9.6       & 86.8       & 82.9 \\
    \bottomrule
\end{tabular}
\caption{\label{table:page1_table} Computation cost (\textbf{G}iga \textbf{FLOPs}), \textbf{Offline F1} and \textbf{Streaming} \textbf{E}xact \textbf{M}atch accuracy of standard bidirectional (BiDi) encoder, \encoder~ model and \name~ (\encoder~ with \adapter~ for guiding Adaptive Restarts) on SNIPS test set.
% The numbers do not reflect state of the art, as these models are smaller and not pretrained.
}
\end{table}

Existing state-of-the-art  bidirectional encoders (such as \newcite{bert_paper}) are computationally expensive for streaming processing. When a new token is received, these models require a \emph{restart}, i.e., re-computation of representation of each token by re-running the bidirectional layer~\cite{towards_incremental_transformers}. This adds to the computational cost (i.e., Higher FLOPs) during streaming and leads to higher wall clock time. Another limitation of these encoders is poorer generalization to partial inputs \cite{incremental_processing_in_the_age_of_non_incremental_encoders}, stemming from these models being trained only on complete (and offline) inputs. Despite these disadvantages, such bidirectional models offer better \emph{offline} performance than unidirectional models across several NLP tasks like sequence tagging \cite{towards_incremental_transformers}.

% \su{this and the next para could be merged. Also, I feel hybrid encoder is a weak contribution to start with. Also refer to the table.}
% \footnote{Restart here refers to computing the bidirectional layer for all tokens from scratch when new tokens arrive as input.}
We address these issues by proposing \name~ – \textbf{H}ybrid \textbf{E}ncoder with \textbf{A}daptive \textbf{R}estarts, where a separate Adaptive Restart Module (\adapter~) predicts when to restart the encoder. The encoder in \name~ is a \encoder~ that reduces the computational cost of running the models in streaming settings. In a \encoder~, the earlier layers are unidirectional and the deeper layers are bidirectional. This design allows early contextualization, and limits the need for restart to the later layers. While \encoder~ reduce streaming computational overhead, restarting them at every new token may not be required. Thus, we propose a lightweight Adaptive Restart Module (\adapter~) to guide restarts, by predicting whether restarting the bidirectional layers of the \encoder~ will benefit the model performance. This module is also compatible with fully bidirectional encoders.

Table \ref{table:page1_table} showcases the strength of \name~ on one of the sequence tagging dataset we consider. In terms of streaming computation, measured in terms of FLOPs (lower is better), \encoder~ offers significant saving from purely bidirectional (BiDi) encoders and FLOPs savings improve upon incorporating \adapter~ in \name~. In terms of offline performance, \encoder~ and \name~ achieves parity with BiDi encoders and in terms of streaming performance, \name~ significantly outperforms BiDi encoders.

Following are our key contributions:

\begin{itemize}
    \item We introduce \encoder~ for computationally cheaper streaming processing (\S\ref{DEBI_subsection}), that maintains the offline F1 score of bidirectional encoders, while reducing FLOPs by an average of 40.2\% across four tasks.
    \item We propose the \adapter~ module (\S\ref{waited_restarts_subsection}) to decide when to restart. The \adapter~ reduces FLOP of \encoder~ by 32.3\% and improves streaming predictions by +4.23 Exact Match.
    \item Our best model \name~ combines \encoder~ with \adapter~ (\S\ref{sec:experimental_results}) to achieve strong streaming performance while saving FLOPs and offering competitive offline performance across four sequence tagging tasks.
    %\item We evaluate sequence tagging in a streaming setting on a wide range of datasets, diverse metrics (\S\ref{sec:experimental_results}) and strong baselines to analyse each component of the model (\S\ref{sec:analysis_ablation}).
    % \item e do ablation}
\end{itemize}

% \citep{Gusfield:97} \citealp{Gusfield:97} \citet{Gusfield:97} \citeyearpar{Gusfield:97}

\section{Streaming Sequence Tagging}

% Here we start by re-introducing the sequence tagging task in the widely assumed offline settings and then move to the streaming setting.

% Consider a sequence tagging task with a token sequence represented as $\bX=(x_1, x_2 \cdots x_n)$ and its ground truth tag sequence $\bY^*=(y^*_1, y^*_2 \cdots y^*_n)$ each of length $n$. In the offline setting, the model receives the entire input text and must predict the offline tag sequence $\hat{\bY} = (\hat{y}_1, \hat{y}_2, \cdots \hat{y}_n)$.

In the streaming sequence tagging task, we assume that at time $1 \le t \le n$, we have received the first $t$ tokens $\bX_{t} = (x_1, \cdots x_t)$ as input from a stream of transcribed speech or user-typed input.\footnote{We assume that new tokens get added without changing previous tokens (contrary to some ASR systems), even though, our method can be used in such settings.} The model then predicts the tags for all $t$ tokens $\hat{\bY}_{t} =(\hat{y}_{1, t}, \hat{y}_{2, t} \cdots \hat{y}_{t, t})$. 
% We assume there to be an end marker to indicate that final input token has been received.
% We also use $x_{<t}$ subscript to denote an arbitrary tokens of prefix that excluding $x_t$. 
% Each $x_{<t}$ is a previous token of $x_t$ and each $x_t$ is a future token for each of $x_{<t}$. 
% At timestep $t$, based on the current token received, the model may update the tag of any of the previous tokens. 
The models predictions $\hat{\bY}_{n}$ over the offline input $\bX = \bX_{n}$ is the offline tag sequence prediction which is evaluated against the ground truth $\bY^*_n=(y^*_1, y^*_2 \cdots y^*_n)$. 
However, in the streaming settings, we are concerned with predicted label sequences over all of the prefixes $\hat{\BY} = (\hat{\bY}_{1}, \hat{\bY}_{2} \cdots \hat{\bY}_{n})$.

During training we only have access to the offline ground-truth label sequence over the offline input sequence $\bY^*_{n}$, even though ground truth labels for tokens may change as additional context is received in the future timesteps. 

%  'BiDi': {'GFLOP': 52.3, 'Streaming EM': 75.1, 'EO': 15.9, 'F1': 93.06},
%  'PBiDi': {'GFLOP': 30.4, 'Streaming EM': 72.5, 'EO': 16.7, 'F1': 93.0},
%  'BiDi+ADAPTER': {'GFLOP': 34.9, 'Streaming EM': 82.8, 'EO': 9.6, 'F1': 93.06},
%  'PEAR': {'GFLOP': 15.1, 'Streaming EM': 85.1, 'EO': 6.3, 'F1': 93.0},
%  'UniDi': {'GFLOP': 9.6, 'Streaming EM': 82.89, 'EO': 0.0, 'F1': 86.76}
% \begin{figure}
% \centering
% \includegraphics[width=\linewidth]{result.pdf}
% \caption{\label{figure:plots} A plot of FLOPs and Prefix exact match across different approaches in streaming settings.}
% \end{figure}

% \input{radar.tex}

\section{Streaming Sequence Tagging with \name~}

In order to motivate \name~, consider how existing BiDi encoders would be used in streaming sequence tagging. At time $t$ and the input sequence $\bX_t$, the model restarts to predicts the label sequence $\hat{\bY}_t$ i.e. it re-computes all its layers for all tokens, without leveraging any of its previous computation or any previously predicted labels. Consequently, running a typical $O(n^2)$ encoder would have $O(n^3)$ computations in streaming settings, over $n$ input tokens. Therefore, naively using existing BiDi encoders for streaming settings is highly inefficient. Previous works tackle inefficiency by modifying the streaming model to infer only once for each word, after $k$ future tokens have been received~\cite{oda_graham_neubig_2015_syntax_based_smt, Graber_Morgan_Daume_2014_rl_for_smt}. This leads to a $k$-delayed output with possibility for revisions after additional future words have been received. However, this may lead to poor performance for tasks involving long-range dependency (e.g., SRL) and a higher output lag.

\begin{figure*}[t!]
\centering
\includegraphics[width=\textwidth]{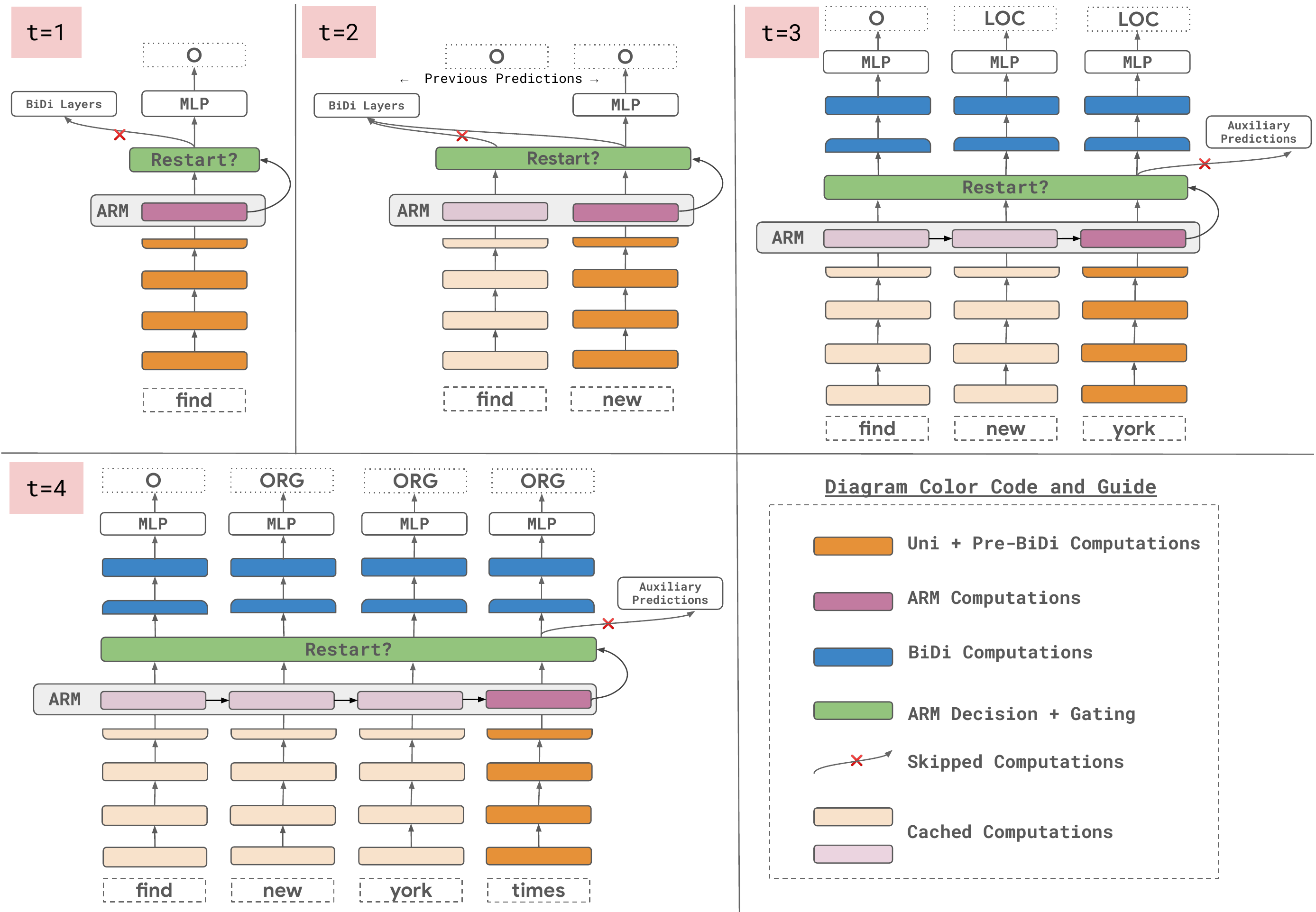}
\caption{\label{figure:approach} 
The \name~ model in streaming setting. At $t=1$, \adapter~ decides not to run the bidirectional layers and exit early through the auxiliary predictor over unidirectional encoding. Similarly at $t=2$, \adapter~ decides not to restart and considers the auxiliary predictions of the latest token and the predictions from the previous steps. At $t=3$ and $t=4$, \adapter~ decides to restart and runs the bidirectional layers to get prediction for all the tokens from scratch. Note that in all the steps, for unidirectional layers and \adapter{}, the computation is performed only for the latest token received.}
\end{figure*}

\name~ is a system consisting of a trained \encoder~ model and an \adapter~ that is trained over the \encoder~. The early unidirectional layers in \encoder~, reduces the computational cost of restart of the encoder. The \adapter~ guides when to restart the \encoder~. It saves computational cost by keeping the model from restarting at each timestep and also improves streaming performance by avoiding unnecessary label flips stemming from avoidable restarts. Figure \ref{figure:approach} shows running of \name~ in streaming settings over an example. For each new token in the streaming input, all the unidirectional layers and a part of first bidirectional layer is ran for the token, to obtain its unidirectional encodings and updated cache. These are then used by \adapter~ to predict whether to restart the bidirectional layer or not. If the bidirectional layer is to be restarted, then we obtain updated labels for all the tokens received in the stream. Otherwise, the auxiliary predictor is ran over the unidirectional encoder for the current token to obtain its label and the labels from previous timestep is copied for all other tokens.% We consider heuristics such as enforcing a maximum and minimum wait for restart over the \adapter~'s predictions. The details of these can be found in the appendix.

We now formally introduce \encoder~ (\S\ref{DEBI_subsection}), followed by \adapter~ (\S\ref{waited_restarts_subsection}).
\subsection{\textsc{Hybrid} Encoder} \label{DEBI_subsection}
\begin{figure}[t]
\centering
\includegraphics[width=\linewidth]{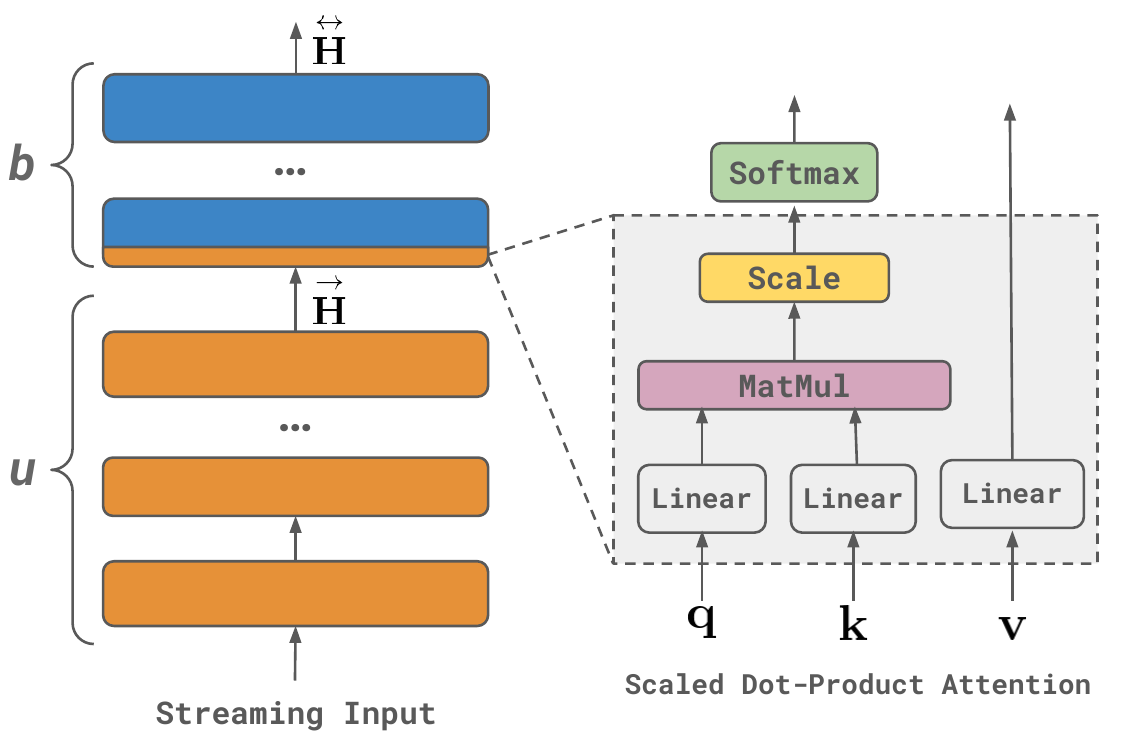}
\caption{\label{figure:hybrid_encoder}In \encoder~ architecture, the earlier $u$ layers are unidirectional, and later $b$ layers are bidirectional. The {\color{diagramblue} blue layers} require restart, i.e. at each timestep when the model receives a new token of the streaming input, these layers will recompute all intermediate representations for all the input tokens. The {\color{orange} orange layers} (like unidirectional layers) can avoid restart by caching the intermediate state. A portion of the first bidirectional layer before application of softmax also does not require restart and can be cached.}
\end{figure}

The \encoder~ is a combination of unidirectional and bidirectional encoding layers, where the early layers are unidirectional and the later layers are bidirectional as shown in Figure \ref{table:encoder_notation}. The earlier unidirectional layers do not require a restart as the previous tokens' embeddings do not need to be updated for the newly received tokens. Thus, the \encoder~ only require a restart for its later bidirectional layers. Formally, the \encoder~ has $u$ unidirectional and $b$ bidirectional layers where total layers in the model are $u + b = l$. Let $\unidirec{L}$ and $\bidirec{L}$ denote unidirectional and bidirectional layers in the model respectively. Each of the $\unidirec{L}$ layers are one of the existing unidirectional layers - RNN, GRU, transformer with causal masking etc. with the first of these being the static embedding layer. Each of the $\bidirec{L}$ layers are bidirectional layers, being one of bi-GRU, bi-LSTM, transformer etc.

\begin{table}[h!]
\small
\setlength{\tabcolsep}{2.5pt}
\centering\begin{tabular}{lp{60mm}}
    \toprule
    \textbf{Notation} & \textbf{Meaning} \\
    \midrule
    $\unidirec{L}$,  $\bidirec{L}$ & Unidirectional and bidirectional layers, respectively. \\
   % $\bidirec{L}$ & Bidirectional Encoding Layers \\
    $\unidirec{\textbf{h}}_i$,  $\bidirec{\textbf{h}}_{i,t}$ & Encoding of $x_i$ token when $t$ tokens have been received, using $\unidirec{L}$ and $\unidirec{L}+\bidirec{L}$, respectively.  \\
   % $\bidirec{\textbf{h}}_{i,j}$ & Bidirectional Encoding of $x_i$ token when $j$ tokens have been received. \\
    $\unidirec{\textbf{H}}_t$, $\bidirec{\textbf{H}}_{t}$ &   $\unidirec{\textbf{h}}_i$,  $\bidirec{\textbf{h}}_{i,t}$ of first $t$ tokens, respectively. \\
    %$\bidirec{\textbf{H}}_{i}$ & Bidirectional Encodings of first $i$ tokens. \\
    $\unidirec{y}_i$,  $\bidirec{y}_{i,t}$ & Predicted label of $x_i$ token using $\unidirec{\textbf{h}}_i$ and $\bidirec{\textbf{h}}_{i,t}$, respectively. \\
   % $\bidirec{y}_{i,j}$ & Predicted label of $x_i$ token using $\bidirec{\textbf{H}}_{i}$ when $j$ tokens have been received. \\
    \bottomrule
\end{tabular}
\caption{\label{table:encoder_notation}  Notations used in \encoder~ for layers, token representations, and predicted labels.}
\end{table}

\subsubsection{\textsc{Hybrid} Encoder in Offline Setting:}

First, the input tokens $\bX$ are fed to $u$ unidirectional layers to get the unidirectional encodings $\unidirec{\bH}$,
\begin{equation}
    \unidirec{\bH} = [\unidirec{\mathbf{h}}_1, \cdots \unidirec{\mathbf{h}}_n] = \overbrace{\unidirec{L} \circ \unidirec{L} \circ \cdots \unidirec{L}}^{u \text{ layers}}(\mathbf{x})
\end{equation}
where $\circ$ denotes the functional composition and $\unidirec{\mathbf{h}}_i$ is the unidirectional encoding for the token $x_i$. Then the bidirectional encodings $\bidirec{\bH}$ are  computed and used to predict the label sequence:
\begin{align}
% \bidirec{\bH} &= [\bidirec{h}_1, \cdots \bidirec{h}_n] \\
% &= \overbrace{\bidirec{L} \circ \bidirec{L} \circ \cdots \bidirec{L}}^{b \text{ layers}}({\unidirec{\bH}})
\bidirec{\bH} = [\bidirec{\mathbf{h}}_1, \cdots \bidirec{\mathbf{h}}_n] = \overbrace{\bidirec{L} \circ \bidirec{L} \circ \cdots \bidirec{L}}^{b \text{ layers}}({\unidirec{\bH}})
\end{align}

The final offline labels are obtained by passing $\bidirec{\bH}$ through a feed-forward neural network layer.

\subsubsection{\textsc{Hybrid} Encoder in Streaming Setting}

In the streaming setting, the unidirectional layers' computation can be cached.\footnote{Cache for RNNs and causally masked transformer consists of hidden states and keys-values respectively. This is similar to implementations of auto-regressive models \cite{huggingface_library, flax_libary}.} These cached intermediate representations for the previously received tokens $\bX_{t-1}$ are used in computing the unidirectional encoding $\unidirec{\mathbf{h}}_t$ of the new token $x_t$. This along with the cached unidirectional encoding of $\bX_{t-1}$ gives us $\unidirec{\bH}_t = [\unidirec{\bH}_{t-1}; \unidirec{\mathbf{h}}_t]$.  
The bidirectional encoding, however, for each token in $\bX_{t}$ is restarted 
at time \emph{t} as $\bidirec{\bH}_t = [\bidirec{\mathbf{h}}_{1, t}, \cdots \bidirec{\mathbf{h}}_{t, t}]$ using the obtained final unidirectional encoding  $\unidirec{\bH}_t$ as input. 

\subsubsection{Training and Inference of \textsc{Hybrid} Encoder}

We predict labels over both $\unidirec{\bH}_t$ and $\bidirec{\bH}_t$ of the \encoder~ using a linear layer with softmax at each timestep. Let $\bidirec{\bY}_{t} = [\bidirec{y}_{1, t}, \cdots \bidirec{y}_{t, t}]$ denote the predictions over bidirectional embeddings at time $t$ for all the tokens received so far. Let $\unidirec{\bY}_t = [\unidirec{y}_1, \cdots \unidirec{y}_t]$ be predictions over unidirectional embeddings at time $t$. While the unidirectional predictions do not perform as well as bidirectional predictions, these auxiliary predictions enables waited restarts (\S\ref{waited_restarts_subsection}).

\encoder~ is trained over the offline input-output sequence pairs and optimize both the bidirectional predictions and unidirectional predictions. for the standard softmax cross entropy loss against the offline ground truth label sequence $\bY^*$.
In order to preserve the strong performance of the BiDi encoders, we inhibit backward flow of gradient from the parameters $\theta_{uni}$ of unidirectional prediction head (consisting only of a linear layer with softmax) to the remaining parameters - $\theta_{bi}$.% \mfar{let's just omit this gradient blocking, it is more likely a bug or an error in optimization and it will confuse the reviewers.}
Following are the losses of these two portions of the model, optimized together with equal weight.\footnote{$CE$ is the Cross Entropy Loss}
\begin{align}
\mathcal{L}(\theta_{bi} , \bidirec{\bY}_n, \bY^*) &= CE(\bidirec{\bY}_{t}, \bY^*) \\
\mathcal{L}(\theta_{uni}, \unidirec{\bY}_n, \bY^*) &= CE(\unidirec{\bY}_{t}, \bY^*) \\
\mathcal{L}(\theta) = \mathcal{L}(\theta_{bi} , \bidirec{\bY}_n, \bY^*) &+ \mathcal{L}(\theta_{uni}, \unidirec{\bY}_n, \bY^*)
\end{align}
% \begin{equation}
% \end{equation}
% All layers are trained end-to-end with backpropagation. 
% We avoid truncated training~\cite{Kohn_2018_incremental_nlp, dalvi_2018_incremental_decoding_training_methods_for_simultaneous_translation_in_nmt}, a commonly used strategy in streaming machine translation as it degrades model performance~\cite{incremental_processing_in_the_age_of_non_incremental_encoders} over streaming sequence labelling. This is because in contrast to machine translation, where techniques like dictionary lookup can approximately the techniques like dictionary lookup for approximately aligning , the sequence labels cannot be approximated from the offline input's labels. \mfar{truncated training won't be known to readers. Does it mean training on partial inputs with gold data dervied from the complete input? Better to just skip truncated training altogether unless it has some consequence later in the paper}
\subsection{Waited Restarts} \label{waited_restarts_subsection}
The \encoder~ reduces the computational overhead of each \emph{restart} by limiting the recomputation over previous tokens only to the later bidirectional layers. However, restarting at each step is not required and the auxiliary predictions from the unidirectional layers can suffice.
% Depending on the current token received, the bidirectional layers of the \encoder~ may update its previous labels or even worsen it for some of the tokens. 
% Thus, restarting at such timesteps not only adds unnecessary computation overhead, but also reduces the prefix performance. 
%To enable this, we introduce mechanisms to prevent the model from restarting at every timestep.

We define a variable $\textsc{restart}(t)$ to decide whether to restart the bidirectional layers at time $t$. When $\textsc{restart}(t)=1$, we restart the model to get the bidirectional predictions $\bidirec{\bY}_{t}$ as the final predicted label sequence $\hat{\bY}_{t} = \bidirec{\bY_{t}}$. If $\textsc{restart}(t)=0$, the model does not run the bidirectional encoder, but uses the unidirectional encoding to predict the current token's label $\unidirec{y}_t$ and copy the label sequence from the previous step $\hat{\bY}_{t-1}$ to obtain $\hat{\bY}_{t}$. Formally, the predicted sequence $\hat{\bY}_{t} = (\hat{y}_{1, t}, \cdots \hat{y}_{t, t})$ at time $t$ is
\begin{equation}\label{waited_restart_equation}
\hat{\bY}_{t} = \begin{cases}
~ \bidirec{\bY}_{t} & \text{if } \textsc{restart}(t)=1 \text{ or } t=n\\
~ [\hat{\bY}_{t-1}; \unidirec{y}_t] & \text{otherwise}
\end{cases}
\end{equation}

where `;' denotes the concatenation. We always restart in the final timestep (i.e., $\textsc{restart}(n) = 1$) to preserve the offline performance.
% and $\mathcal{C}$ is the conditional function that picks $\bidirec{\bY}_{t}$ if $\textsc{wait}(t)$, a selector binary variable, is True otherwise picks $[\hat{\bY}^{(t-1)}; \unidirec{y}_t]$. 

% Setting $\textsc{wait}(t)$ to 0 (1) at each timestep, then the bidirectional layer will never (always) be restarted.
\subsubsection{Adaptive Restart Module (ARM)}
 A baseline restart strategy for waited restarts would be to restart the bidirectional layers every fixed $k$ steps, i.e., $\textsc{restart}(t)=1$ whenever $t$ is a multiple of $k$. Note that $k=1$ reduces to a \encoder~ model without waited restarts. We refer to this as \textbf{\textsc{Restart-$k$}}.
 
Rather than having a heuristic function for $\textsc{restart}(t)$, we can let a lightweight parametric module determine when to do the restart by determining ${\textsc{restart}}_\theta{(t)}$ for each $t$. We refer to this as \textbf{A}daptive \textbf{R}estart \textbf{M}odule (\textbf{ARM}). We first discuss the set of features for \adapter~ (\S\ref{subsubsection:features_for_adapter}), followed by its architecture (\S\ref{subsubsection:adapter_architecture}) and training (\S\ref{subsubsection:adaptor_training}).

\subsubsection{Input Features for \adapter~}\label{subsubsection:features_for_adapter}
We use \encoder~'s intermediate representations that do not require restart as the features for \adapter~. This includes the unidirectional layers as well as from the pre-softmax features from the first $\bidirec{L}$ layer. The main motivation to use these features is to incorporate more information without incurring any restarts.

The following features are stored in unidirectional cache and used without any restart for \adapter{}: 
\begin{enumerate}[label=(\roman*)]
\item \emph{unidirectional encodings} $\unidirec{\mathbf{h}}_t$ from the last unidirectional layer computed once and cached for capturing the backward flow,
\item \emph{query} $\mathbf{q}_t$ and \emph{key} $\mathbf{k}_t$ from the first bidirectional layer computed only from $\unidirec{\mathbf{h}}_t$,
\item \emph{unnormalized forward attention scores} $\mathbf{a}_t = [\mathbf{q}_1 \cdot \mathbf{k}_t, \cdots, \mathbf{q}_{t-1} \cdot \mathbf{k}_t]$ from the first bidirectional layer, where `$\cdot$' is the dot product. These scores are concatenated across the prefix tokens (restricted and padded to latest $m$ tokens) across all heads. 
\end{enumerate}
These features are then concatenated and provided as input to the \adapter~ as $\mathbf{f}_t = [\unidirec{\mathbf{h}}_t, \mathbf{q}_t, \mathbf{k}_t, \mathbf{a}_t]$.
%unidirectional encoding, query and keys in first bidirectional layer and unnormalized attention scores can be stored in the unidirectional cache.

\subsubsection{\adapter~ Architecture}\label{subsubsection:adapter_architecture}
   
We use a single-layered $\textsc{Gru}$ \cite{gru_paper} to predict restart. The probability of restart $p_\theta(\textsc{restart}(t)=1)$ is modeled as,
\begin{align}
(\mathbf{o}_t, \mathbf{s}_t) &= \text{\textsc{Gru}}(\mathbf{f}_t, \mathbf{s}_{t-1}) \\
p_\theta(\textsc{restart}(t)=1) &= \sigma(\textsc{Linear}(\mathbf{o}_t))
\end{align}
Here $\mathbf{s}_t$ and $\mathbf{o}_t$ are the GRU hidden state and output at time $t$, respectively. 
These outputs can be post-processed to handle too frequent or too infrequent waiting and obtain the final selector predictions.% \mfar{This isn't clear. Are you saying you can have different threshold values on p(restart) to decide to restart or not. Just say it here. Assume reader won't open the appendix.}
Details can be found in Appendix (\S\ref{appendix:subsection:postprocessing_adapter}).

\subsubsection{Training the \adapter~} \label{subsubsection:adaptor_training}

Ideally, \encoder~ should restart only when it would lead to improved predictions over the prefixes i.e., more number of prefix inputs should have the same output as the the offline ground truth for those prefixes. We use this to define our ground truth \textsc{restart} sequence for training the \adapter~.  

Given the ground truth offline tag sequence $\bY^*$, unidirectional predictions $\unidirec{\bY} = [\unidirec{y}_1, \cdots \unidirec{y}_n]$ and final bidirectional prediction sequence over each of the $n$ prefixes $\bidirec{\BY} = [\bidirec{\bY}_{1}, \cdots \bidirec{\bY}_{n}]$, we define the ground truth policy $\textsc{Restart}^*(t) = \pi^*_t$ for \adapter~, at time $t$, as follows.

\begin{equation}\label{waited_restart_equation}
 \pi^*_t = \begin{cases}
 1 & \text{if } |y^*_{\le t} = \unidirec{y}_{\le t}| < |y^*_{\le t} = \bidirec{y}_{\le t}|\\
 0 & \text{otherwise}
\end{cases}
\end{equation}

The policy decides to restart at time $t$ if more tokens in bidirectional predictions match with the ground truth ($|y^*_{\le t} = \bidirec{y}_{\le t}|$) than those from unidirectional predictions ($|y^*_{\le t} = \unidirec{y}_{\le t}|$).\footnote{This policy is greedy. The optimal policy obtained via dynamic programming relies on the future tokens.}
\adapter~ is trained against this policy with features from a frozen and trained \encoder~ model with a standard binary cross entropy loss.

\begin{table*}[t!]
\small
\setlength{\tabcolsep}{2.7pt}
\centering\begin{tabularx}{\textwidth}{Y   l  | Y | Y | Y | Y }
    \toprule
    Current Token & Prefix Predictions & Total & Unnecessary & Exact Match & Exact Match  \\
     &  & Edits & Edits & w/ ground truth & w/ final prediction \\ \midrule
    find   & \texttt{O}                & 1 & 0 & 1 & 1  \\
    new    & \texttt{O O}                    & 2 & 1 & 1 & 1  \\
    york   & \texttt{O LOC LOC}              & 4 & 3 & 2 & 1  \\
    times  & \texttt{O ORG ORG ORG}           & 7 & 3 & 2 & 2  \\
    square & \texttt{O ORG ORG ORG LOC}       & 8 & 3 & 2 & 3  \\
    \midrule
    Ground Truth & \texttt{O LOC LOC LOC LOC} & \multicolumn{4}{c}{\tiny \textbf{EO}=$ \frac{\text{\tiny Unnecessary ~Edits}}{ \text{\tiny Total ~Edits}} = \frac{3}{8}$  ~~ \textbf{Streaming EM} = $\frac{\text{\tiny EM w/ ground truth}}{ \text{\tiny \# steps}} =\frac{2}{5}$ ~~ \textbf{RC} = $\frac{\text{\tiny EM w/ final labels}}{ \text{\tiny \# steps}} =\frac{3}{5}$} \\
    \bottomrule
\end{tabularx}
\caption{\label{table:metrics_streaming_example} Example computation for the metrics described in \S\ref{metrics} . \textbf{EO} is calculated as fraction of edits that were \emph{unnecessary}. \textbf{RC} measures the fraction of prefix predictions that matches the model's final predictions. \textbf{Streaming EM} measures the fraction of prefix predictions that are correct with respect to the ground truth labels.}
\end{table*}
\section{Experimental Setup}

\paragraph{Datasets and Tasks.}
\label{datasets}
We consider four common sequence tagging tasks -- Slot Filling over SNIPS~\cite{snipsdataset}, Semantic Role Labelling (SRL) over Ontonotes~\cite{ontonotes_dataset}, and Named Entity Recognition (NER) and Chunking (CHUNK) over CoNLL-2003~\cite{conll2003dataset}. The standard train, dev and test splits are used for all datasets.

\paragraph{Models and Baselines.}

For our models, we consider a layer budget of 4 for the encoder. Naturally, our baselines are the UniDi encoder and BiDi encoder with all unidirectional and bidirectional layers, respectively. For \name~, we tune for the optimal fractions of unidirectional layers in encoder by selecting the one that maximizes the offline F1 performance over the development set. We consider \textsc{Restart}-$k$ baseline for \adapter~. For each dataset, we picked the best value of $k$ from $\{2, 3, 5, 8\}$, maximizing the Streaming EM over development set.

\subsection{Metrics}

\label{metrics}

The model evaluation is done over three criteria: offline performance over complete input text, streaming performance over prefixes and efficiency of running the model in streaming settings.

\paragraph{Offline Metrics.} The offline performance of the model is measured using the widely used chunk-level F1 score for sequence labelling tasks~\cite{conll2003dataset}. 
%Higher offline F1 score is better.

\paragraph{GFLOPs.} We measure total number of Floating Point Operations (FLOPs) required for running the model in a streaming setting in GigaFLOPs (\textbf{GFLOP}), estimated via XLA compiler's  High Level Operations (HLO) \cite{xla_paper}. GFLOP positively correlates with how computation-heavy a model is and its wall-clock execution time.

\paragraph{Streaming Metrics.} An ideal streaming model should predict the correct labels for all prefixes early \cite{trinh_2018_multi_task_inc_state_tracking} and avoid unnecessary label flips. Following previous works \cite{incremental_processing_in_the_age_of_non_incremental_encoders, towards_incremental_transformers}, we use Edit Overhead (EO) and Relative Correctness (RC) metrics from \newcite{baumann_2011_evaluation_and_optimization_of_incremental_processor}.

% \ag{Restructured and paraphrased a few things}
EO (Edit Overhead) is a measure of the fraction of the label edits that were unnecessary with respect to the final prediction on complete output. Consider the example in Table \ref{table:metrics_streaming_example}. At the first timestep, the token ``\emph{find}'' is assigned a label ``\texttt{O}'' from ``\texttt{N/A}''; taking the total edits to $1$. Similarly, in the second timestep, the newly received token ``\textit{new}'' gets a label edit; taking the total edit to $2$. In the third timestep, however, not only the newly received token ``\textit{york}'' receives a label edit but the second token ``\textit{new}'' is also edited from ``\texttt{O}'' to ``\texttt{LOC}''; resulting in $4$ total edits. Of these edits, the label edit for token "\textit{new}" in timestep 2 from ``\texttt{N/A}'' to ``\texttt{O}'' was unnecessary as it differs from its final label ``\texttt{ORG}''. Similarly, the label flips for ``\textit{new}'' and ``\textit{york}'' in the timestep 3 from ``\texttt{O}'' and ``\texttt{N/A}'' to ``\texttt{LOC}'' and ``\texttt{LOC}'', respectively, were also unnecessary. Cumulatively, towards the end of fifth timestep, we have $3$ out of $8$ total edits which were \emph{overhead}. Thus, the EO turns out to be $\frac{3}{8}$, where lower is better.

RC measures the relative correctness of prefixes, i.e. correctness of prefix prediction sequence with respect to the final label set over the complete input. For the example in Table \ref{table:metrics_streaming_example}, only the label sequence in the first timestep (``\texttt{O}''), fourth timestep (``\texttt{O ORG ORG ORG}'') and final (fifth) timestep are prefixes of the label sequence in the final timestep (``\texttt{O ORG ORG ORG LOC}''). Thus $3$ prefixes of a total $5$ prefixes were correct. So, the Relative Correctness is $\frac{3}{5}$, where higher is better.

While EO and RC capture consistency and stability in streaming predictions, neither of these measures performance with respect to the ground truth label over offline input. Relying on these metrics alone is not sufficient to measure streaming performance. For example, a UniDi encoder achieves perfect EO and RC scores, despite have poor predictions with respect to the offline ground truth. Thus, we consider \textbf{Streaming E}xact \textbf{M}atch (\textbf{Streaming EM}), which is the streaming setting analogue of the Exact Match metric. Streaming EM calculates the percentages of prefix label sequence which are correct with respect to the offline ground truth labels. For example in Table \ref{table:metrics_streaming_example}, only the first label sequence (``\texttt{O}'') for input ``\textit{find}'' and third label sequence (``\texttt{O LOC LOC}'') for input ``\textit{find new york}'' is a prefix of the ground truth label sequence (``\texttt{O LOC LOC LOC LOC}''), leading to only $2$ out of $5$ prefix label sequence being correct. Thus, the Streaming EM is $\frac{2}{5}$. Similar to RC, higher score is better as more prefixes have exact matches.
\begin{table}[t!]
\small
\setlength{\tabcolsep}{4pt}
\centering\begin{tabular}{c c c c c}
    \toprule
    Model        & SNIPS         & CHUNK            & NER              & SRL \\
    \midrule
    UniDi encoder       & 86.8          & 88.1             & 73.8             & 56.4 \\
    \encoder~    & 93.0          & \textbf{89.4}    & \textbf{86.8}    & 80.0 \\
    BiDi  encoder       & \textbf{93.1} & \textbf{89.4}    & 86.0             & \textbf{80.1} \\
    \bottomrule
\end{tabular}
\caption{\label{table:conll_f1_DEBI} Offline F1 over test set of datasets described in \S\ref{datasets}. On all datasets, \encoder~ achieves performance parity with the BiDi models.}
\end{table}
\begin{table*}
\small
\centering
\setlength{\tabcolsep}{4.5pt}
\begin{tabular}{l | c c c | c c c | c c c | c c c }
 \toprule

 Dataset & \multicolumn{3}{c|}{GFLOP $\downarrow$} & \multicolumn{3}{c|}{Streaming EM $\uparrow$} & \multicolumn{3}{c|}{EO $\downarrow$} &
 \multicolumn{3}{c}{RC $\uparrow$}\\
 \midrule
 & BiDi & \textsc{Hybrid} & \name~ & BiDi & \textsc{Hybrid} & \name~ & BiDi & \textsc{Hybrid} & \name~ & BiDi & \textsc{Hybrid} & \name~ \\
 \midrule
 % RC
 % Bidi 78.21 91.62 87.21 55.96
 % Hybrid 76.31 91.87 88.15 61.46
 % HEAR 88.53 91.58 90.85 62.45
 SNIPS  & 74.7 & 43.4 & \textbf{21.6}
        & 75.1  & 72.5  & \textbf{85.1}
        & 15.9  & 16.7  & \textbf{6.3} 
        & 78.2  & 76.3  & \textbf{88.5} \\
 CHUNK  & 238.8 & 126.3 & \textbf{91.3}
        & 77.7  & 77.7  & \textbf{77.9}
        & 5.0   & 4.8   & \textbf{4.5}
        & 91.6  & \textbf{91.9}  & 91.6 \\
 NER    & 238.8 & 126.2 & \textbf{83.8}
        & 79.3  & 78.2  & \textbf{81.9}
        & 8.7   & 8.3   & \textbf{5.4}
        & 87.2  & 88.2  & \textbf{90.9} \\
 SRL    & 741.2 & 557.7 & \textbf{460.4}
        & 43.6  & 49.1  & \textbf{50.6}
        & 33.0  & 29.9  & \textbf{21.0}
        & 56.0  & 61.5 & \textbf{62.5} \\
 \bottomrule
\end{tabular}
\caption{\label{table:streaming_flops_trc_eo} Streaming performance over test set of datasets. \name~ significantly outperforms BiDi on all metrics improving upon efficiency and performance. Direction of arrow indicates whether higher or lower is better. }
\end{table*}

\section{Experimental Results}\label{sec:experimental_results}
In this section, we provide empirical results to answer the following questions: 
% \begin{enumerate}[label=(\alph*)] %% replace enumerate* with enumerate for vertical list
%     \item In offline setting, does \encoder~ achieve parity of a fully bidirectional Models (\S\ref{subsec:conll_f1_DEBI})?
%     \item In a streaming setting, does \encoder~ reduce streaming GFLOP count while maintaining offline F1 score (\S\ref{subsec:GFLOPs_DEBI})?
%     \item In a streaming setting, does the \adapter~ reduce GFLOP count and improve streaming performance (\S\ref{subsec:restarts})? 
%     \item In a streaming setting, how does \name~ model (\encoder~ + \adapter~) compares to the vanilla BiDi model  (\S\ref{sec:final_model_eval_comparison})?
% \end{enumerate}

\begin{enumerate}[label=(\alph*)] %% replace enumerate* with enumerate for vertical list
    \item In offline setting, does \encoder~ achieve parity with BiDi encoders?
    \item In streaming setting, by how much does \encoder~ reduce GFLOPs count?% \mfar{This has an obvious answer - yes. I think the right question is: Can hybrid encoder achieve quality parity with bidirectional encoder in streaming setting while reducing GFLOP count?}
    \item Does \name~ improve streaming performance and save on GFLOP count?
    % \item In a streaming setting, how does \name~ model (\encoder~ + \adapter~) compares to the vanilla BiDi model (\S\ref{sec:final_model_eval_comparison})?
\end{enumerate}

% \subsection{Offline Setting Evaluation of \encoder~}\label{subsec:conll_f1_DEBI}
\paragraph{\textsc{Hybrid} Encoder's Offline Performance is Competitive to BiDi Encoders.} Table \ref{table:conll_f1_DEBI} shows the offline F1 scores of the \encoder~, UniDi and BiDi encoders across the four tasks. The \encoder~ has similar offline F1 as BiDi encoders. In fact, on NER, it even outperforms it by a margin of 0.8 F1 score. As expected, the UniDi model performs poorly compared to the BiDi encoder across all the tasks. From here on, we omit UniDi encoder. 

\paragraph{\textsc{Hybrid} Encoder Improves Streaming Efficiency.} Table \ref{table:streaming_flops_trc_eo} shows the GFLOP (per input instance) in streaming settings across the four datasets for the \encoder~ and BiDi encoders, with a trivial restart at every new token to get predictions over streaming text. We observe that \encoder~ consistently offers lower GFLOP count than BiDi encoder across the datasets, and in three of the four datasets, offering more than 40\% FLOP reduction. However, \encoder~ does not improve on streaming performance (Streaming EM, EO, and RC) over BiDi, and its FLOP can be further reduced. We next see how incorporating \name~ addresses this.

\paragraph{\name~ Improves Streaming Efficiency and Performance.} Across all datasets, we observe \name~ further reduces GFLOP count from the already computationally lighter \encoder~, giving us upto 71\% total FLOP reduction from BiDi. \name~ has better streaming predictions, as its Streaming EM is much higher (upto +10.0) than BiDi, improving on the shortcoming of Hybrid. This highlights that naively restarting at each timestep can worsen the streaming performance, as evident through the lower performance of Hybrid and BiDi. \name~ also offers reduced number of avoidable label flips, as signified by a lower EO. Its high RC score signifies that its streaming predictions are more consistent with its final prediction.

All these results demonstrate that \name~ preserves the offline performance of BiDi encoders, while being computationally lighter by 58.9\% on average across tasks. \name~ has more consistent streaming predictions which are accurate with respect to offline ground truth.

\section{Analysis and Ablations}
\label{sec:analysis_ablation}

\begin{table*}
\small
\centering
\begin{tabular}{l | c c | c c | c c | c c }
 \toprule

 Dataset & \multicolumn{2}{c|}{Offline F1$\uparrow$} & \multicolumn{2}{c|}{Streaming EM $\uparrow$} & \multicolumn{2}{c|}{EO $\downarrow$} & \adapter~ Classification \\
 \midrule
 & BiDi & UniDi & \textsc{restart}-$k$ & \adapter~ & \textsc{restart}-$k$ & \adapter~ & Micro-F1 \\
 \midrule
 SNIPS  & \textbf{93.6}  & 79.1
        & 82.1  & \textbf{85.3} & 7.9 & \textbf{6.0}
        & 79.5 \\
 CHUNK  & \textbf{90.4}  & 86.8
        & 75.6  & \textbf{76.5} & 4.5 & \textbf{4.5}
        & 67.1 \\
 NER    & \textbf{91.3}  & 77.9
        & 85.6  & \textbf{87.8} & 4.1 & \textbf{4.0}
        & 75.7 \\
 SRL    & \textbf{79.7}  & 42.8
        & 41.8 & \textbf{48.9} & 31.5 & \textbf{21.6}
        & 80.2 \\
 \bottomrule
\end{tabular}
\caption{\label{table:analysis_ablations_combined} Development set ablation and analyses. From left to right - comparing performance of \encoder~ over sequence tagging using its bidirectional (final) vs unidirectional (intermediate) encoding, comparison of Streaming EM and EO of \textsc{restart}-k vs \adapter~, and the performance of \adapter~ w.r.t.  its ground truth policy labels.}
\end{table*}

In this section, we perform various analysis pertaining to the \encoder~ and \adapter~.

% \begin{enumerate}[label=(\alph*)] %% replace enumerate* with enumerate for vertical list
%     \item Performance of \encoder~'s unidirectional encodings (\S\ref{subsection:encoder_unidirectional_performance})?
%     \item Comparison of \adapter~ to \textsc{Restart}-$k$ (\S\ref{subsection:adaptive_vs_fixed_restart}).
%     \item Intrinsic performance of \adapter~ (\S\ref{subsection:intrinsic_adapter_performance}).
%     % \item \adapter~'s architecture ablation (\S\ref{subsection:adapter_architecture_ablation})?

% \end{enumerate}

\subsection{\textsc{Hybrid} Encoder's Unidirectional Layers Performance} \label{subsection:encoder_unidirectional_performance}

% \begin{table}[h!]
% \small
% \centering
% \begin{tabular}{l c c c c }
%  \toprule
%  Dataset & SNIPS & CHUNK & NER & SRL \\
%  \midrule
%  Final        & 93.6 & 90.4 & 91.3 & 79.7 \\
%  Intermediate & 79.1 & 86.8 & 77.9 & 42.8 \\
%  \bottomrule
% \end{tabular}
% \caption{\label{table:unidirectional_performance} Dev set offline F1 score for the best \name~ performance using final bidirectional features after bidirectional vs intermediate unidirectional features.
% \end{table}

Table \ref{table:analysis_ablations_combined} compares the performance of the prediction over the intermediate (unidirectional) and final (bidirectional) encodings for the best \name~ model. While the intermediate ones lag in comparison to the final, we get decent performance from intermediate ones across all except for SRL dataset. This shows that, when used selectively, UniDi intermediate predictions can serve as a good source of auxiliary predictions.

\subsection{\adapter~ vs \textsc{Restart}-$k$.} \label{subsection:adaptive_vs_fixed_restart}
% \ak{use the consistent term for fixed restart}

% \begin{table}[h!]
% \small
% \centering
% \begin{tabular}{l  c c  c c}
%  \toprule
%  Dataset & \multicolumn{2}{c}{\textsc{Restart}-$k$} & \multicolumn{2}{c}{\adapter} \\
%  \midrule
%   & Streaming EM $\uparrow$ & EO $\downarrow$ & Streaming EM $\uparrow$ & EO $\downarrow$ \\
%  \midrule
%  SNIPS & 82.1 & 7.9  & \textbf{85.3} & \textbf{6.0} \\
%  CHUNK & 75.6 & 4.5  & \textbf{76.5} & \textbf{4.5} \\
%  NER   & 85.6 & 4.1  & \textbf{87.8} & \textbf{4.0} \\
%  SRL   & 41.8 & 31.5 & \textbf{48.9} & \textbf{21.6} \\
%  \bottomrule
% \end{tabular}
% \caption{\label{table:wait-k_adapter} Streaming EM (Higher is better) and EO (lower is better) over dev set.}
% \end{table}

Table \ref{table:analysis_ablations_combined} shows the performance of baseline - the best \textsc{Restart}-$k$ from \{2, 3, 5, 8\} over dev set Streaming EM
and the \adapter~. We observe that across all the datasets, on both Streaming EM and EO metric, \adapter~ performs significantly better than the heuristic \textsc{Restart}-$k$ mechanism. This shows that using a fixed length wait is not sufficient.

\subsection{Intrinsic Performance of \adapter~} 
\label{subsection:intrinsic_adapter_performance}

% \begin{table}[h!]
% \small
% \centering
% \begin{tabular}{l c c c c }
%  \toprule
%  Dataset & SNIPS & CHUNK & NER & SRL \\
%  \midrule
%  Micro-F1 & 79.5 & 67.1 & 75.7 & 80.2 \\
%  \bottomrule
% \end{tabular}
% \caption{\label{table:adapter_performance} \adapter~ classification performance over dev set. }
% \end{table}

Table \ref{table:analysis_ablations_combined} shows the performance of the \adapter~ on its binary classification task. Given the lightweight \adapter~ architecture and the task complexity, the performance is satisfactory with 80.2 $F_1$. However, there is a considerable margin for its improvement both in terms of model architecture and features.

\subsection{\adapter~'s Architecture Ablation} \label{subsection:adapter_architecture_ablation}

\begin{table}[h!]
\small
\centering
\begin{tabular}{l c c }
 \toprule
 Dataset & Streaming EM $\uparrow$ & EO $\downarrow$ \\
 \midrule

 No \adapter~     & 73.8 & 16.3 \\
 Linear \adapter~ & 81.3 & 6.8  \\
 MLP \adapter~    & 81.4 & 6.9  \\
 GRU \adapter~    & \textbf{85.3} & \textbf{6.0} \\
 \bottomrule
\end{tabular}
\caption{\label{table:arm_model_ablation} Performance of \name~ with different \adapter~ model architectures over SNIPS development set.}
\end{table}

Table \ref{table:arm_model_ablation} shows the Streaming EM and EO scores of \name~ with different \adapter~ model architectures for the SNIPS development set. We observe that \name~ having either a Linear layer or MLP as \adapter~ does much better than having no \adapter~ and restarting at each timestep. However, modeling \adapter~ as a GRU recurrent model gives the best scores in both metrics.

\section{Related Work}

Streaming (or incremental) setting has been widely studied in machine translation and parsing, dating as far back as two decades \cite{larchev_optimal_incremental_parsing_1995, lane_henderson_2001_incremental_syntactic_parsing}. Specifically, for incremental parsing, there are two broad approaches that have been studied: transition-based and graph-based. Transition-based incremental parsers allow for limited backtracking and correcting parsing over prefixes by keeping track of multiple parse segments \cite{buckman-etal-2016-transition} or via beams search \cite{bhargava-penn-2020-supertagging}. Such methods can fail on garden-path sentences and long-range dependencies and only work proficiently with a large beam. Graph-based parsers incrementally assign scores to edges of the graph, discarding those edges that cause conflicts to tree-structure of the graph. \cite{yang_deng_2020_strongly_incremental_constituency_parsing_with_gnn} proposed an attach-juxtapose system to grow the tree, requiring restart at each new token over streaming input. \cite{parsing_acl22_best_paper} improved on its efficiency by proposing a information bottleneck. However, these methods rely on the structured nature of parsing output can not be adapted to incremental sequence tagging tasks without restarting at each token.
% More recently, \cite{dary-nasr-2021-reading} extended incremental parsing techniques to tasks like tokenization, PoS Tagging. None of these works, however, focus on making streaming inference efficient.

Recently, ~\newcite{incremental_processing_in_the_age_of_non_incremental_encoders} benchmarked the modern encoders on streaming sequence labelling and observed poor streaming performance of pretrained transformer models. They explored improvement strategies by adopting techniques from other streaming streaming like chunked training ~\cite{dalvi_2018_incremental_decoding_training_methods_for_simultaneous_translation_in_nmt}, truncated training ~\cite{Kohn_2018_incremental_nlp} (training model on heuristically-aligned partial input-output pairs), and prophecy~\cite{prediction_of_future_words_smt} (autocompleting input using a separate language model). They observed performance degradation from truncated training unless used with prophecies. However, running a language model at each timestep for prophecy is computationally infeasible. Therefore, unlike our approach, neither methods can improve performance feasibly.

Previous works have attempted to improve computation efficiency in BiDi encoders. Monotonic attention moves away from $O(n^2)$ soft attention overhead by restricting attention to monotonically increase across timesteps~\cite{monotonic_attention, monotonic_chunkwise_attention, monotonic_multiheaded_attention}. However, unlike \name~, such attempts can't maintain offline performance. Recently, \newcite{towards_incremental_transformers} used linear transformer ~\cite{linear_transformer_paper} as unidirectional model using masking for streaming sequence tagging and classification. This approach performs well under the assumption of delayed output ~\cite{Graber_Morgan_Daume_2014_rl_for_smt, oda_graham_neubig_2015_syntax_based_smt}, a relaxation, where the model waits for additional tokens before predicting. Furthermore, the unidirectional model could not revise its output, rendering the model incapable of handling long-range dependency or tasks that go backward like SRL — an ability common to any model with some bidirectionality, such as \name~. Similar drawbacks were in the partial bidirectional encoder, a bidirectional attention with restricted window, proposed by \newcite{from_simultaneous_to_streaming_machine_translation}.

Improving efficiency through adaptive computing has been independently studied for reasoning-based tasks~ \cite{adaptive_computing_multimodal_learning}, text generation models~ \cite{calm_adaptive_computing, dact_bert_adaptive} and diffusion models~ \cite{diffusion_models_variable_timesteps_Sept_2022}. These works are restricted to offline settings and can be readily incorporated within the proposed overall approach of \name{}.

\section{Conclusion}

We propose \name~ for sequence tagging in streaming setting where the input is received one token at a time to the model. The encoder in our model is \encoder~ where early layers are unidirectional and later are bidirectional. It reduces the computational cost in streaming settings, by reducing the need of restart only to the later bidirectional layers while preserving the offline performance of the model. \name~ additionally consists of an \adapter~ to predict when to restart the model. Using \adapter~ leads to reduced number of restart of the encoder, leading to better streaming performance and further savings in computation. Compared to BiDi encoders, our model, \name~ (\encoder~ + \adapter~) reduces the computation by upto 71\% in streaming settings while maintaining the performance of the BiDi encoders across various sequence tagging tasks. \name~ improved streaming EM by upto +10.0\% and reduced unnecessary edits by upto -12.0\%.

\section*{Limitations}

An additional but small training cycle is required to train the lightweight \adapter~ module of \name~ in order to reap the benefits of extra savings in efficiency and streaming performance.
Also, even though we do not assume any language specific-design choices, we benchmarked on the standard streaming sequence labelling benchmark datasets, all of which were in English.
% Our work focuses on making the streaming NLU systems less computation heavy. Therefore dampening the environmental implications of running neural models.
% Multilingual.

% Delayed Output, Truncated Training and Prophecies, pretraining can be incorporated

\bibliography{anthology,custom}

\begin{thebibliography}{42}
\expandafter\ifx\csname natexlab\endcsname\relax\def\natexlab#1{#1}\fi

\bibitem[{Alinejad et~al.(2018)Alinejad, Siahbani, and
  Sarkar}]{prediction_of_future_words_smt}
Ashkan Alinejad, Maryam Siahbani, and Anoop Sarkar. 2018.
\newblock \href {https://doi.org/10.18653/v1/D18-1337} {Prediction improves
  simultaneous neural machine translation}.
\newblock In \emph{Proceedings of the 2018 Conference on Empirical Methods in
  Natural Language Processing}, pages 3022--3027, Brussels, Belgium.
  Association for Computational Linguistics.

\bibitem[{Arumae and Bhatia(2020)}]{calm_adaptive_computing}
Kristjan Arumae and Parminder Bhatia. 2020.
\newblock \href {http://arxiv.org/abs/arXiv:2004.03794} {Calm: Continuous
  adaptive learning for language modeling}.

\bibitem[{Baumann et~al.(2011)Baumann, Buß, and
  Schlangen}]{baumann_2011_evaluation_and_optimization_of_incremental_processor}
Timo Baumann, Okko Buß, and David Schlangen. 2011.
\newblock \href {https://doi.org/https://doi.org/10.5087/dad.2011.106}
  {Evaluation and optimisation of incremental processors}.
\newblock In \emph{Dialogue and Discourse, 2011}.

\bibitem[{Bhargava and Penn(2020)}]{bhargava-penn-2020-supertagging}
Aditya Bhargava and Gerald Penn. 2020.
\newblock \href {https://doi.org/10.18653/v1/2020.repl4nlp-1.23} {Supertagging
  with {CCG} primitives}.
\newblock In \emph{Proceedings of the 5th Workshop on Representation Learning
  for NLP}, pages 194--204, Online. Association for Computational Linguistics.

\bibitem[{Bradbury et~al.(2018)Bradbury, Frostig, Hawkins, Johnson, Leary,
  Maclaurin, Necula, Paszke, Vander{P}las, Wanderman-{M}ilne, and
  Zhang}]{jax_paper}
James Bradbury, Roy Frostig, Peter Hawkins, Matthew~James Johnson, Chris Leary,
  Dougal Maclaurin, George Necula, Adam Paszke, Jake Vander{P}las, Skye
  Wanderman-{M}ilne, and Qiao Zhang. 2018.
\newblock \href {http://github.com/google/jax} {{JAX}: composable
  transformations of {P}ython+{N}um{P}y programs}.

\bibitem[{Buckman et~al.(2016)Buckman, Ballesteros, and
  Dyer}]{buckman-etal-2016-transition}
Jacob Buckman, Miguel Ballesteros, and Chris Dyer. 2016.
\newblock \href {https://doi.org/10.18653/v1/D16-1254} {Transition-based
  dependency parsing with heuristic backtracking}.
\newblock In \emph{Proceedings of the 2016 Conference on Empirical Methods in
  Natural Language Processing}, pages 2313--2318, Austin, Texas. Association
  for Computational Linguistics.

\bibitem[{Cai and de~Rijke(2016)}]{query_auto_completion}
Fei Cai and Maarten de~Rijke. 2016.
\newblock \href {https://doi.org/10.1561/1500000055} {A survey of query auto
  completion in information retrieval}.
\newblock \emph{Found. Trends Inf. Retr.}, 10(4):273–363.

\bibitem[{Chang et~al.(2022)Chang, Prakash, Wu, Liang, Sainath, Li, Stambler,
  Upadhyay, Faruqui, and Strohman}]{streaming_query_processing}
Shuo-yiin Chang, Guru Prakash, Zelin Wu, Qiao Liang, Tara~N. Sainath, Bo~Li,
  Adam Stambler, Shyam Upadhyay, Manaal Faruqui, and Trevor Strohman. 2022.
\newblock \href {http://arxiv.org/abs/arXiv:2208.13322} {{Streaming Intended
  Query Detection using E2E Modeling for Continued Conversation}}.

\bibitem[{Chiu and Raffel(2018)}]{monotonic_chunkwise_attention}
Chung-Cheng Chiu and Colin Raffel. 2018.
\newblock \href {https://openreview.net/forum?id=Hko85plCW} {Monotonic
  chunkwise attention}.
\newblock In \emph{International Conference on Learning Representations}.

\bibitem[{Cho and Esipova(2016)}]{neural_smt_2016}
Kyunghyun Cho and Masha Esipova. 2016.
\newblock \href {http://arxiv.org/abs/arXiv:1606.02012} {Can neural machine
  translation do simultaneous translation?}

\bibitem[{Cho et~al.(2014)Cho, van Merrienboer, Gulcehre, Bahdanau, Bougares,
  Schwenk, and Bengio}]{gru_paper}
Kyunghyun Cho, Bart van Merrienboer, Caglar Gulcehre, Dzmitry Bahdanau, Fethi
  Bougares, Holger Schwenk, and Yoshua Bengio. 2014.
\newblock \href {http://arxiv.org/abs/arXiv:1406.1078} {{Learning Phrase
  Representations using RNN Encoder-Decoder for Statistical Machine
  Translation}}.

\bibitem[{Coucke et~al.(2018)Coucke, Saade, Ball, Bluche, Caulier, Leroy,
  Doumouro, Gisselbrecht, Caltagirone, Lavril, Primet, and
  Dureau}]{snipsdataset}
Alice Coucke, Alaa Saade, Adrien Ball, Théodore Bluche, Alexandre Caulier,
  David Leroy, Clément Doumouro, Thibault Gisselbrecht, Francesco Caltagirone,
  Thibaut Lavril, Maël Primet, and Joseph Dureau. 2018.
\newblock \href {http://arxiv.org/abs/arXiv:1805.10190} {{Snips Voice Platform:
  an Embedded Spoken Language Understanding system for private-by-design voice
  interfaces}}.

\bibitem[{Dalvi et~al.(2018)Dalvi, Durrani, Sajjad, and
  Vogel}]{dalvi_2018_incremental_decoding_training_methods_for_simultaneous_translation_in_nmt}
Fahim Dalvi, Nadir Durrani, Hassan Sajjad, and Stephan Vogel. 2018.
\newblock \href {https://doi.org/10.18653/v1/N18-2079} {Incremental decoding
  and training methods for simultaneous translation in neural machine
  translation}.
\newblock In \emph{Proceedings of the 2018 Conference of the North {A}merican
  Chapter of the Association for Computational Linguistics: Human Language
  Technologies, Volume 2 (Short Papers)}, pages 493--499, New Orleans,
  Louisiana. Association for Computational Linguistics.

\bibitem[{Devlin et~al.(2019)Devlin, Chang, Lee, and Toutanova}]{bert_paper}
Jacob Devlin, Ming-Wei Chang, Kenton Lee, and Kristina Toutanova. 2019.
\newblock \href {https://doi.org/10.18653/v1/N19-1423} {{{BERT}: Pre-training
  of Deep Bidirectional Transformers for Language Understanding}}.
\newblock In \emph{Proceedings of the 2019 Conference of the North {A}merican
  Chapter of the Association for Computational Linguistics: Human Language
  Technologies, Volume 1 (Long and Short Papers)}, Minneapolis, Minnesota.
  Association for Computational Linguistics.

\bibitem[{Eyzaguirre et~al.(2022)Eyzaguirre, del Rio, Araujo, and
  Soto}]{dact_bert_adaptive}
Cristobal Eyzaguirre, Felipe del Rio, Vladimir Araujo, and Alvaro Soto. 2022.
\newblock \href {https://doi.org/10.18653/v1/2022.nlppower-1.10}
  {{DACT}-{BERT}: Differentiable adaptive computation time for an efficient
  {BERT} inference}.
\newblock In \emph{Proceedings of NLP Power! The First Workshop on Efficient
  Benchmarking in NLP}, pages 93--99, Dublin, Ireland. Association for
  Computational Linguistics.

\bibitem[{Eyzaguirre and Soto(2020)}]{adaptive_computing_multimodal_learning}
Cristóbal Eyzaguirre and Álvaro Soto. 2020.
\newblock \href {https://doi.org/10.1109/CVPR42600.2020.01283} {Differentiable
  adaptive computation time for visual reasoning}.
\newblock In \emph{2020 IEEE/CVF Conference on Computer Vision and Pattern
  Recognition (CVPR)}, pages 12814--12822.

\bibitem[{Grissom~II et~al.(2014)Grissom~II, He, Boyd-Graber, Morgan, and
  Daum{\'e}~III}]{Graber_Morgan_Daume_2014_rl_for_smt}
Alvin Grissom~II, He~He, Jordan Boyd-Graber, John Morgan, and Hal
  Daum{\'e}~III. 2014.
\newblock \href {https://doi.org/10.3115/v1/D14-1140} {Don{'}t until the final
  verb wait: Reinforcement learning for simultaneous machine translation}.
\newblock In \emph{Proceedings of the 2014 Conference on Empirical Methods in
  Natural Language Processing ({EMNLP})}, pages 1342--1352, Doha, Qatar.
  Association for Computational Linguistics.

\bibitem[{Gu et~al.(2017)Gu, Neubig, Cho, and Li}]{neural_smt_2017}
Jiatao Gu, Graham Neubig, Kyunghyun Cho, and Victor~O.K. Li. 2017.
\newblock \href {https://aclanthology.org/E17-1099} {{Learning to Translate in
  Real-time with Neural Machine Translation}}.
\newblock In \emph{Proceedings of the 15th Conference of the {E}uropean Chapter
  of the Association for Computational Linguistics: Volume 1, Long Papers},
  pages 1053--1062, Valencia, Spain. Association for Computational Linguistics.

\bibitem[{He et~al.(2017)He, Lee, Lewis, and
  Zettlemoyer}]{luheng_he_lee_lewis_zettlemoyer_2017_deep_acl}
Luheng He, Kenton Lee, Mike Lewis, and Luke Zettlemoyer. 2017.
\newblock \href {https://doi.org/10.18653/v1/P17-1044} {{Deep Semantic Role
  Labeling: What Works and What{'}s Next}}.
\newblock In \emph{Proceedings of the 55th Annual Meeting of the Association
  for Computational Linguistics (Volume 1: Long Papers)}, pages 473--483,
  Vancouver, Canada. Association for Computational Linguistics.

\bibitem[{Heek et~al.(2020)Heek, Levskaya, Oliver, Ritter, Rondepierre,
  Steiner, and van {Z}ee}]{flax_libary}
Jonathan Heek, Anselm Levskaya, Avital Oliver, Marvin Ritter, Bertrand
  Rondepierre, Andreas Steiner, and Marc van {Z}ee. 2020.
\newblock \href {http://github.com/google/flax} {{F}lax: A neural network
  library and ecosystem for {JAX}}.

\bibitem[{Iranzo~Sanchez et~al.(2022)Iranzo~Sanchez, Civera, and
  Juan-C{\'\i}scar}]{from_simultaneous_to_streaming_machine_translation}
Javier Iranzo~Sanchez, Jorge Civera, and Alfons Juan-C{\'\i}scar. 2022.
\newblock \href {https://doi.org/10.18653/v1/2022.acl-long.480} {From
  simultaneous to streaming machine translation by leveraging streaming
  history}.
\newblock In \emph{Proceedings of the 60th Annual Meeting of the Association
  for Computational Linguistics (Volume 1: Long Papers)}, pages 6972--6985,
  Dublin, Ireland. Association for Computational Linguistics.

\bibitem[{Kahardipraja et~al.(2021)Kahardipraja, Madureira, and
  Schlangen}]{towards_incremental_transformers}
Patrick Kahardipraja, Brielen Madureira, and David Schlangen. 2021.
\newblock \href {https://doi.org/10.18653/v1/2021.emnlp-main.90} {{Towards
  Incremental Transformers: An Empirical Analysis of Transformer Models for
  Incremental {NLU}}}.
\newblock In \emph{Proceedings of the 2021 Conference on Empirical Methods in
  Natural Language Processing}, pages 1178--1189, Online and Punta Cana,
  Dominican Republic. Association for Computational Linguistics.

\bibitem[{Katharopoulos et~al.(2020)Katharopoulos, Vyas, Pappas, and
  Fleuret}]{linear_transformer_paper}
Angelos Katharopoulos, Apoorv Vyas, Nikolaos Pappas, and Fran{\c{c}}ois
  Fleuret. 2020.
\newblock \href {https://proceedings.mlr.press/v119/katharopoulos20a.html}
  {Transformers are {RNN}s: Fast autoregressive transformers with linear
  attention}.
\newblock In \emph{Proceedings of the 37th International Conference on Machine
  Learning}, volume 119 of \emph{Proceedings of Machine Learning Research},
  pages 5156--5165. PMLR.

\bibitem[{Kingma and Ba(2014)}]{adam_paper}
Diederik~P. Kingma and Jimmy Ba. 2014.
\newblock \href {http://arxiv.org/abs/arXiv:1412.6980} {{Adam: A Method for
  Stochastic Optimization}}.

\bibitem[{Kitaev et~al.(2022)Kitaev, Lu, and Klein}]{parsing_acl22_best_paper}
Nikita Kitaev, Thomas Lu, and Dan Klein. 2022.
\newblock \href {https://doi.org/10.18653/v1/2022.acl-long.220} {Learned
  incremental representations for parsing}.
\newblock In \emph{Proceedings of the 60th Annual Meeting of the Association
  for Computational Linguistics (Volume 1: Long Papers)}, pages 3086--3095,
  Dublin, Ireland. Association for Computational Linguistics.

\bibitem[{K{\"o}hn(2018)}]{Kohn_2018_incremental_nlp}
Arne K{\"o}hn. 2018.
\newblock \href {https://aclanthology.org/C18-1253} {Incremental natural
  language processing: Challenges, strategies, and evaluation}.
\newblock In \emph{Proceedings of the 27th International Conference on
  Computational Linguistics}, pages 2990--3003, Santa Fe, New Mexico, USA.
  Association for Computational Linguistics.

\bibitem[{Lane and
  Henderson(2001)}]{lane_henderson_2001_incremental_syntactic_parsing}
Peter C.~R. Lane and James~B. Henderson. 2001.
\newblock Incremental syntactic parsing of natural language corpora with simple
  synchrony networks.
\newblock \emph{IEEE Transactions on Knowledge and Data Engineering}, 13:2001.

\bibitem[{Larchev\^{e}que(1995)}]{larchev_optimal_incremental_parsing_1995}
J.-M. Larchev\^{e}que. 1995.
\newblock \href {https://doi.org/10.1145/200994.200996} {Optimal incremental
  parsing}.
\newblock \emph{ACM Trans. Program. Lang. Syst.}, 17(1):1–15.

\bibitem[{Ma et~al.(2020)Ma, Pino, Cross, Puzon, and
  Gu}]{monotonic_multiheaded_attention}
Xutai Ma, Juan~Miguel Pino, James Cross, Liezl Puzon, and Jiatao Gu. 2020.
\newblock \href {https://openreview.net/forum?id=Hyg96gBKPS} {Monotonic
  multihead attention}.
\newblock In \emph{International Conference on Learning Representations}.

\bibitem[{Madureira and
  Schlangen(2020)}]{incremental_processing_in_the_age_of_non_incremental_encoders}
Brielen Madureira and David Schlangen. 2020.
\newblock \href {https://doi.org/10.18653/v1/2020.emnlp-main.26} {{Incremental
  Processing in the Age of Non-Incremental Encoders: An Empirical Assessment of
  Bidirectional Models for Incremental {NLU}}}.
\newblock In \emph{Proceedings of the 2020 Conference on Empirical Methods in
  Natural Language Processing (EMNLP)}, pages 357--374, Online. Association for
  Computational Linguistics.

\bibitem[{Oda et~al.(2015)Oda, Neubig, Sakti, Toda, and
  Nakamura}]{oda_graham_neubig_2015_syntax_based_smt}
Yusuke Oda, Graham Neubig, Sakriani Sakti, Tomoki Toda, and Satoshi Nakamura.
  2015.
\newblock \href {https://doi.org/10.3115/v1/P15-1020} {Syntax-based
  simultaneous translation through prediction of unseen syntactic
  constituents}.
\newblock In \emph{Proceedings of the 53rd Annual Meeting of the Association
  for Computational Linguistics and the 7th International Joint Conference on
  Natural Language Processing (Volume 1: Long Papers)}, pages 198--207,
  Beijing, China. Association for Computational Linguistics.

\bibitem[{Pennington et~al.(2014)Pennington, Socher, and Manning}]{glove_paper}
Jeffrey Pennington, Richard Socher, and Christopher Manning. 2014.
\newblock \href {https://doi.org/10.3115/v1/D14-1162} {{G}lo{V}e: Global
  vectors for word representation}.
\newblock In \emph{Proceedings of the 2014 Conference on Empirical Methods in
  Natural Language Processing ({EMNLP})}, pages 1532--1543, Doha, Qatar.
  Association for Computational Linguistics.

\bibitem[{Pradhan et~al.(2013)Pradhan, Moschitti, Xue, Ng, Bj{\"o}rkelund,
  Uryupina, Zhang, and Zhong}]{ontonotes_dataset}
Sameer Pradhan, Alessandro Moschitti, Nianwen Xue, Hwee~Tou Ng, Anders
  Bj{\"o}rkelund, Olga Uryupina, Yuchen Zhang, and Zhi Zhong. 2013.
\newblock \href {https://aclanthology.org/W13-3516} {Towards robust linguistic
  analysis using {O}nto{N}otes}.
\newblock In \emph{Proceedings of the Seventeenth Conference on Computational
  Natural Language Learning}, pages 143--152, Sofia, Bulgaria. Association for
  Computational Linguistics.

\bibitem[{Raffel et~al.(2017)Raffel, Luong, Liu, Weiss, and
  Eck}]{monotonic_attention}
Colin Raffel, Minh-Thang Luong, Peter~J. Liu, Ron~J. Weiss, and Douglas Eck.
  2017.
\newblock \href {https://proceedings.mlr.press/v70/raffel17a.html} {Online and
  linear-time attention by enforcing monotonic alignments}.
\newblock In \emph{Proceedings of the 34th International Conference on Machine
  Learning}, volume~70 of \emph{Proceedings of Machine Learning Research},
  pages 2837--2846. PMLR.

\bibitem[{Ratinov and Roth(2009)}]{lev_ratinov_dan_roth_2009_design}
Lev Ratinov and Dan Roth. 2009.
\newblock \href {https://aclanthology.org/W09-1119} {{Design Challenges and
  Misconceptions in Named Entity Recognition}}.
\newblock In \emph{Proceedings of the Thirteenth Conference on Computational
  Natural Language Learning ({C}o{NLL}-2009)}, pages 147--155, Boulder,
  Colorado. Association for Computational Linguistics.

\bibitem[{Sabne(2020)}]{xla_paper}
Amit Sabne. 2020.
\newblock Xla : Compiling machine learning for peak performance.

\bibitem[{Sang and Meulder(2003)}]{conll2003dataset}
Erik F. Tjong~Kim Sang and Fien~De Meulder. 2003.
\newblock \href {https://aclanthology.org/W03-0419} {{Introduction to the
  {C}o{NLL}-2003 Shared Task: Language-Independent Named Entity Recognition}}.
\newblock In \emph{Proceedings of the Seventh Conference on Natural Language
  Learning at {HLT}-{NAACL} 2003}, pages 142--147.

\bibitem[{Trinh et~al.(2018)Trinh, Ross, and
  Kelleher}]{trinh_2018_multi_task_inc_state_tracking}
Anh~Duong Trinh, Robert~J. Ross, and John~D. Kelleher. 2018.
\newblock \href {https://doi.org/doi:10.21427/cvkg-0p89} {{A Multi-Task
  Approach to Incremental Dialogue State Tracking}}.
\newblock In \emph{SEMDIAL 2018 (AixDial): the 22nd workshop on the Semantics
  and Pragmatics of Dialogue}, Aix-en-Provence, France.

\bibitem[{Wolf et~al.(2020)Wolf, Debut, Sanh, Chaumond, Delangue, Moi, Cistac,
  Rault, Louf, Funtowicz, Davison, Shleifer, von Platen, Ma, Jernite, Plu, Xu,
  Le~Scao, Gugger, Drame, Lhoest, and Rush}]{huggingface_library}
Thomas Wolf, Lysandre Debut, Victor Sanh, Julien Chaumond, Clement Delangue,
  Anthony Moi, Pierric Cistac, Tim Rault, Remi Louf, Morgan Funtowicz, Joe
  Davison, Sam Shleifer, Patrick von Platen, Clara Ma, Yacine Jernite, Julien
  Plu, Canwen Xu, Teven Le~Scao, Sylvain Gugger, Mariama Drame, Quentin Lhoest,
  and Alexander Rush. 2020.
\newblock \href {https://doi.org/10.18653/v1/2020.emnlp-demos.6}
  {{Transformers: State-of-the-Art Natural Language Processing}}.
\newblock In \emph{Proceedings of the 2020 Conference on Empirical Methods in
  Natural Language Processing: System Demonstrations}, pages 38--45, Online.
  Association for Computational Linguistics.

\bibitem[{Yang and
  Deng(2020)}]{yang_deng_2020_strongly_incremental_constituency_parsing_with_gnn}
Kaiyu Yang and Jia Deng. 2020.
\newblock \href
  {https://proceedings.neurips.cc/paper/2020/file/f7177163c833dff4b38fc8d2872f1ec6-Paper.pdf}
  {Strongly incremental constituency parsing with graph neural networks}.
\newblock In \emph{Advances in Neural Information Processing Systems},
  volume~33, pages 21687--21698. Curran Associates, Inc.

\bibitem[{Ye et~al.(2022)Ye, Wu, and
  Liu}]{diffusion_models_variable_timesteps_Sept_2022}
Mao Ye, Lemeng Wu, and Qiang Liu. 2022.
\newblock \href {http://arxiv.org/abs/arXiv:2209.01170} {First hitting
  diffusion models}.

\bibitem[{Zhou et~al.(2022)Zhou, Eisner, Newman, Platanios, and
  Thomson}]{online_semantic_parsing_acl22}
Jiawei Zhou, Jason Eisner, Michael Newman, Emmanouil~Antonios Platanios, and
  Sam Thomson. 2022.
\newblock \href {https://doi.org/10.18653/v1/2022.acl-long.110} {Online
  semantic parsing for latency reduction in task-oriented dialogue}.
\newblock In \emph{Proceedings of the 60th Annual Meeting of the Association
  for Computational Linguistics (Volume 1: Long Papers)}, pages 1554--1576,
  Dublin, Ireland. Association for Computational Linguistics.

\end{thebibliography}
\bibliographystyle{acl_natbib}

\appendix

\section{Experiment Details}
\subsection{Implementation Details}
All the experiments were done using the Flax~\cite{flax_libary} and Jax~\cite{jax_paper} with Adam optimizer~\cite{adam_paper} on TPUs. We measure flops in terms of XLA's HLO \cite{xla_paper}. The optimizer's learning rate is set to 1e-3, betas are set to default at (0.9, 0.999), batch size is 256 and feedforward hiden is 2048 with 512 transformer dimension and 8 heads.

For all the datasets we use the standard splits, the links to which can be found in their respective papers along with statistics. Following \cite{lev_ratinov_dan_roth_2009_design}, we use the BIOES tagging scheme for the NER task and BIO scheme for the rest. In SRL, following \cite{luheng_he_lee_lewis_zettlemoyer_2017_deep_acl} the indicator for predicate verb is also used as input the along with the sentence. 

All the values over test sets are averaged across four seeds. For experiments involving \adapter~ we average across four different \encoder~ trained from different seeded initialization. We initialize static embeddings with 300 dimensional glove embedding \cite{glove_paper}.

\subsection{Postprocessing \adapter~ Predictions}\label{appendix:subsection:postprocessing_adapter}

We post process the \adapter~ predictions to prevent too frequent or too infrequent restarts. Specifically, for hyperparameters $\alpha$ and $\beta$ where $0 \le \alpha < \beta$, at any time $t$, if the \adapter~ hasn't restarted once since $max(0, t-\beta)$ timesteps, then \adapter~'s prediction is set to 1, else, if the \adapter~ has restarted atleast once since $max(0, t-\alpha)$ timesteps, then it is set to 0. We tune for the values of $\alpha$ and $\beta$ in the range \{0, 1, 2, 3, 5, 10\} over the development set.

We also observe that even if bidirectional layers improve predictions over previous tokens, for the latest token only, the unidirectional label $\unidirec{y}_t$ is often better than $\bidirec{y}_t$. If this is observed, then, we exclude the most recent token from getting updated during restart. We tune for this binary postprocessor over the development set as well.

\section{Analysis and Ablation}

\subsection{Benefits of Optimizing Unidirectional Predictor Separately} \label{subsec:stopgrad}

In \encoder~ the performance we inhibit backward gradient flow from the auxiliary predictor over the unidirectional layers unidirectional. This is because we observed offline F1 performance drop of bidirectional predictor when trained with unidirectional predictor, dropping from 81.20 to 71.62 on CoNLL development set. Even after tuning for loss scaling for the to predictors in ratio \{1:1, 3.3:1, 10:1, 33:1, 100:1\}, the performance was only increased upto 75.06 F1.

% \begin{table}[h!]
% \small
% \centering
% \begin{tabular}{| c | c c | c c |}
%  \toprule
%  Dataset & \multicolumn{2}{c|}{with Stop Gradient} & \multicolumn{2}{c|}{without Stop Gradient} \\
%   & Aux & Final & Aux & Final \\
%  \midrule
%  SNIPS & 79.1 & 93.6 &  \\
%  CHUNK & 86.8 & 90.4 & 0 \\
%  NER   & 91.3 & 77.3 & 0 \\
%  SRL   & 42.8 & 79.7 & 0 \\
%  \bottomrule
% \end{tabular}
% \caption{\label{table:stop_gradient} EO over test set. Lower is better. EO for unidirectional models is 0 as they cannot update the labels from previous timesteps.}
% \end{table}

\subsection{Oracle Policy for \adapter~} \label{subsec:policy}

We also experimented with alternate policy for \adapter~ where it's label is conditioned on the last bidirectional restart. However, it led to a performance drop in terms of Streaming EM from 86.7 to 83.0, as observed across SNIPS development set. We attribute this poor performance on alternate policy, due to the lightweight nature of \adapter~. Thus, we did not proceed with this policy.

% \begin{table}[h!]
% \small
% \centering
% \begin{tabular}{l c c }
%  \toprule
%  Dataset & Streaming EM $\uparrow$ & EO $\downarrow$ \\
%  \midrule
%  Greedy Policy    & 86.7 & 2.3 \\
%  Alternate Policy & 83.0 & 0.7 \\
%  \bottomrule
% \end{tabular}
% \caption{\label{table:arm_policy_ablation} Performance of \name~ with different \adapter~ policy over SNIPS development set.}
% \end{table}

\end{document}